\documentclass[11pt,a4paper]{article}
\usepackage[hyperref]{acl2021}
\usepackage{times}
\usepackage{latexsym}

\usepackage{microtype}

\usepackage{booktabs}
\usepackage{caption}
\usepackage{graphicx}
\usepackage{amsmath}
\usepackage{amsfonts}
\usepackage{siunitx}
\usepackage{subcaption}
\usepackage{placeins}

\aclfinalcopy 

\title{Unsupervised Cross-Domain Prerequisite Chain Learning \\using Variational Graph Autoencoders}

\author{Irene Li, Vanessa Yan, Tianxiao Li, Rihao Qu \and Dragomir Radev \\
Yale University, USA \\
\{irene.li,vanessa.yan,tianxiao.li,rihao.qu,dragomir.radev\}@yale.edu\\
}

\date{}

\begin{document}
\maketitle
\begin{abstract}

Learning prerequisite chains is an essential task for efficiently acquiring knowledge in both known and unknown domains. For example, one may be an expert in the natural language processing (NLP) domain but want to determine the best order to learn new concepts in an unfamiliar Computer Vision domain (CV). Both domains share some common concepts, such as machine learning basics and deep learning models. In this paper, we propose unsupervised cross-domain concept prerequisite chain learning using an optimized variational graph autoencoder. Our model learns to transfer concept prerequisite relations from an information-rich domain (source domain) to an information-poor domain (target domain), substantially surpassing other baseline models. Also, we expand an existing dataset by introducing two new domains––CV and Bioinformatics (BIO). The annotated data and resources, as well as the code, will be made publicly available.

\end{abstract}

\section{Introduction}

With the rapid growth of online educational resources in diverse fields, people need an efficient way to acquire new knowledge. Building a concept graph can help people design a correct and efficient study path \cite{alsaad2018mining,yu2020mooccube}. There are mainly two approaches to learning prerequisite relations between concepts: one is to extract the relations directly from course content, video sequences, textbooks, or Wikipedia articles \cite{yang2015concept,pan-etal-2017-prerequisite,alzetta2019prerequisite}, but this approach requires extra work on feature engineering and keyword extraction. Our method follows a different approach of inferring the relations within a concept graph \cite{liang2018investigating,Li2019WhatSI,li-etal-2020-r}.

\begin{figure}[t]
    \centering
    \includegraphics[width=7.5cm]{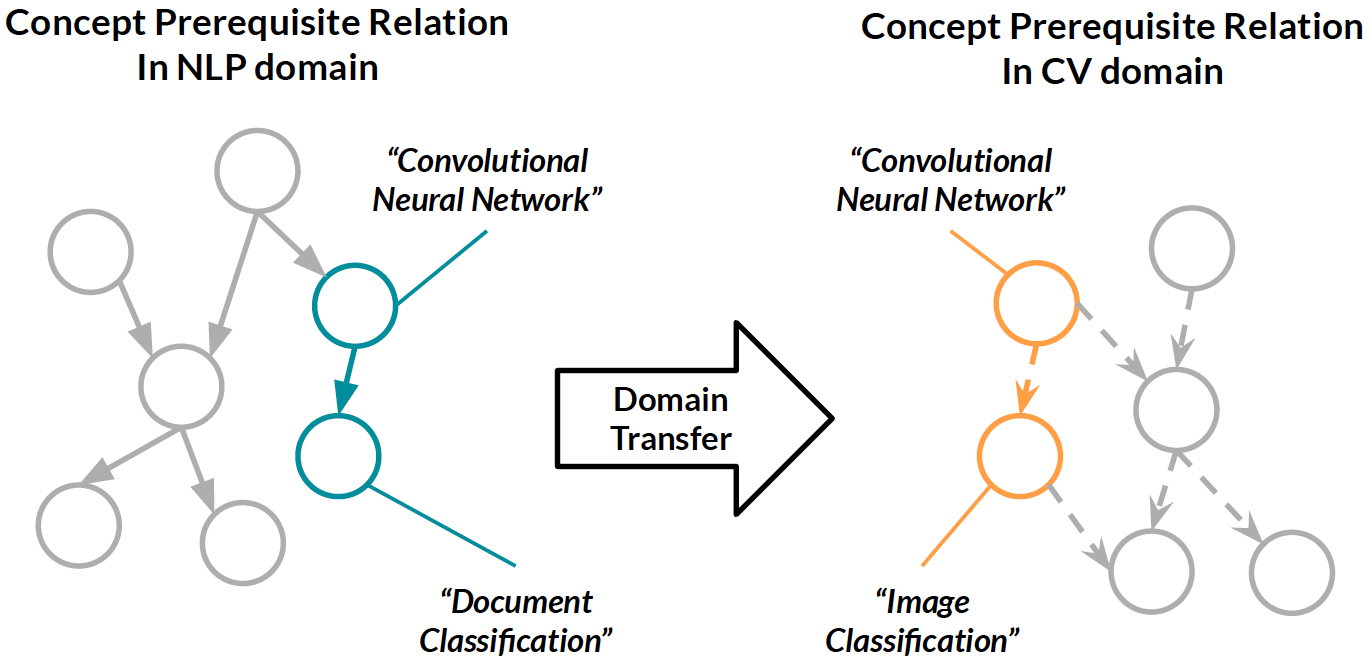}
    \caption{Cross-domain prerequisite chains.}
    \label{fig:intro}
\end{figure}


In a concept graph, we define $p\to q$ as the notion that learning concept $p$ is a prerequisite to learning concept $q$. Existing methods formulate this question as a classification task. A typical method is to encode concept pairs and train a classifier to predict if there is a prerequisite relation \cite{alzetta2019prerequisite,yu2020mooccube}. However, this method requires annotated prerequisite pairs during training. Alternatively, others have used graph-based models to predict prerequisite relations. \citet{gordon-etal-2016-modeling} proposed information-theoretic approaches to infer concept dependencies. \citet{Li2019WhatSI} modeled a concept graph using Variational Graph Autoencoders (VGAE) \cite{kipf2016variational}, training their model to infer unseen prerequisite relations in a semi-supervised way. While most of the previous methods were supervised or semi-supervised, \citet{li-etal-2020-r} introduced Relational-VGAE, which enabled unsupervised learning on prerequisite relations.

Existing work mainly focuses on prerequisite relations within a single domain. In this paper, we tackle the task of cross-domain prerequisite chain learning, by transferring prerequisite relations between concepts from a relatively information-rich domain (source domain) to an information-poor domain (target domain). As an example, we illustrate in Figure \ref{fig:intro}, a partial concept graph from the Natural Language Processing (NLP) domain and a partial concept graph from the Computer Vision (CV) domain. Prerequisite relations among concepts in the NLP domain are known, and we seek to infer prerequisite relations among concepts in the CV domain. These two domains share some concepts, such as \textit{Convolutional Neural Network}. We assume that being aware of prerequisite relations among concepts in the source domain helps infer potential relations in the target domain. More specifically, in the figure, knowing that \textit{Convolutional Neural Network}$\to$\textit{Document Classification} helps us determine that \textit{Convolutional Neural Network}$\to$\textit{Image Classification}. 

Our contributions are two-fold. First, we propose cross-domain variational graph autoencoders to perform unsupervised prerequisite chain learning in a heterogeneous graph. Our model is the first to do domain transfer within a single graph, to the best of our knowledge. Second, we extend an existing dataset by collecting and annotating resources and concepts in two new target domains. Data and code will be made public in \url{https://github.com/Yale-LILY/LectureBank/tree/master/LectureBankCD}.



\begin{table}[b]
\centering
\small
\begin{tabular}{llllll} \toprule
Domain & Files & Pages & Tks/pg & Con. & PosRel  \\\midrule
NLP &    1,717 &  65,028 & 47 & 322 &  1,551 \\
 CV      &  1,041     &   58,32  & 43  &  201        &  871   \\
 BIO     &  148     &  7,13  & 135  &   100       & 234    \\
\bottomrule
\end{tabular}
\caption{LectureBankCD statistics on NLP, CV and BIO domain: Tks/pg (Tokens per slide page), Con. (Number of concepts), PosRel (Positive Relations).}
\label{tab:stats}

\end{table}

\section{Dataset}
LectureBank2.0 \cite{li-etal-2020-r} dataset contains 1,717 lecture slides (hereon called \textbf{resources}) and 322 concepts with annotated prerequisite relations, largely from NLP. We treat this dataset as our information-rich source domain (NLP). Also, we propose an expansion dataset, LectureBankCD, by introducing two new target domains in the same data format: CV and Bioinformatics (BIO). We report statistics on the dataset in Table \ref{tab:stats}. For each domain, we identify high-quality lecture slides from the top university courses, collected by strong background domain experts. And choose concepts by crowd-sourcing. We end up with 201 CV concepts and 100 BIO concepts.  In each domain, we ask two graduate-level annotators with deep domain knowledge to add prerequisite chain annotations for every possible pair of concepts. The Cohen's kappa agreement scores \cite{mchugh2012interrater} are 0.6396 for CV and 0.8038 for BIO. Cohen's kappa between 0.61–0.80 is considered substantial, so our annotations are reliable. We take the union of the positive annotations for our experiments: 871 positive relations for CV and 234 positive relations for BIO.


\section{Methodology}

\begin{figure}[t]
    \centering
    \includegraphics[width=0.44\textwidth]{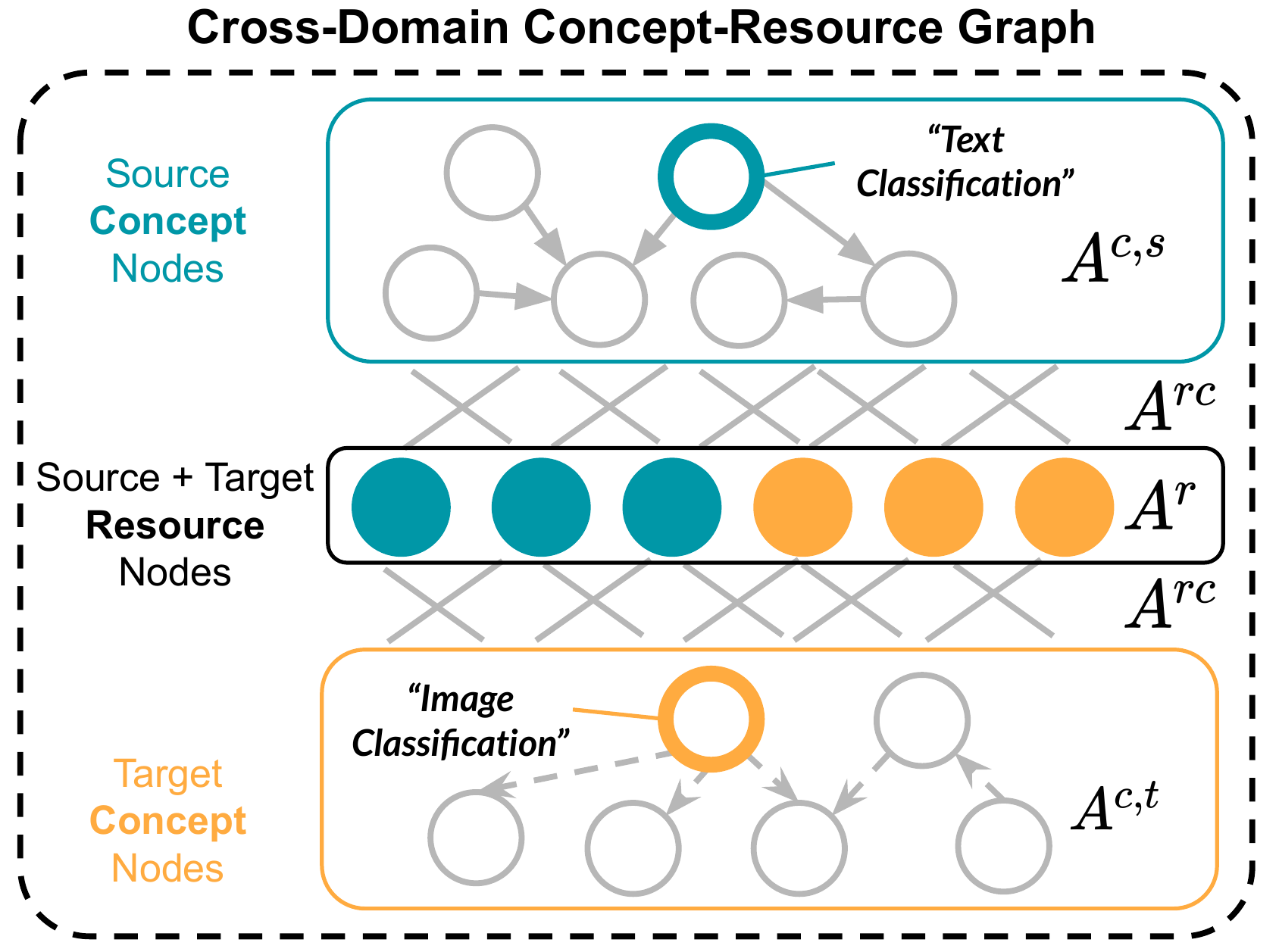}
    \caption{Cross-Domain Concept-Resource Graph: we model the resource nodes (solid nodes) and concept nodes (hollow nodes) from two domains (in blue and orange) in a heterogeneous graph. We show a subset of nodes and edges.}
    \label{fig:model}
\end{figure}

Inspired by \newcite{li-etal-2020-r}, we build a cross-domain concept-resource graph $G=(X,A)$ that includes resource nodes and concept nodes from both the source and target domains (Figure \ref{fig:model}). To obtain the node feature matrix $X$, we use either BERT \cite{devlin2019bert} or Phrase2Vec \cite{artetxe2018emnlp} embeddings.
We consider four edge types to build the adjacency matrix $A$: $A^{c,s}$: edges between source concept nodes; $A^{rc}$: edges between all resource nodes and concept nodes; $A^{r}$: edges between resource nodes only; and $A^{c,t}$: edges between target concept nodes.
In unsupervised prerequisite chain learning, $A^{c,s}$\textemdash concept relations of the source domain\textemdash are known, and the task is to predict $A^{c,t}$\textemdash concept relations of the target domain. For $A^{rc}$ and $A^{r}$, we calculate cosine similarities based on node embeddings, consistent with previous works \cite{Li2019WhatSI,chiu2020autoencoding}.

\textbf{Cross-Domain Graph Encoder} 
VGAE \cite{kipf2016variational} contains a graph neural network (GCN) encoder \cite{kipf2016semi} and an inner product decoder. In a GCN, the hidden representation of a node $i$ in the next layer is computed using only the information of direct neighbours and the node itself. To account for cross-domain knowledge, we additionally
consider the \textit{domain neighbours} for each node $i$. These domain neighbours are a set of common or semantically similar concepts from the other domain.\footnote{In Figure \ref{fig:model}, the two labeled nodes are domain neighbors.} We define the cross-domain graph encoder as:
\begin{equation}
\nonumber
\begin{split}
h_{ i }^{ (l+1) }&=\sigma \Bigg( \Bigg.\sum _{ j\in { N }_{ i } } W^{ (l) }h_{ j }^{ (l) } + W^{ (l) }h_{ i }^{ (l) } \\
&+\sum _{ k\in { N }^{D}_{ i } } W_{D}^{ (l) }h_{ k }^{ (l) } \Bigg.\Bigg)  
\end{split}
\label{eq:cdgcn}
\end{equation}

where $N_{i}$ denotes the set of direct neighbours of node $i$, $N_{i}^{D}$ is the set of domain neighbours, and $W_D$ and $W$ are trainable weight matrices. To determine the domain neighbors, we compute cosine similarities and match the concept nodes only from source domain to target domain: $cosine(h_s,h_t)$. The values are then normalized into the range of [0,1], and we keep top 10\% of domain neighbors.\footnote{Parameter is selected using validation dataset.}  

\textbf{DistMult Decoder}
We optimize the original inner product decoder from VGAE. To predict the link between a concept pair $(c_i,c_j)$, we apply the DistMult \cite{yang2014embedding} method: we take the output node features from the last layer, $\hat{X}$, and define the following score function to recover the adjacency matrix $\hat{A}$ by learning a trainable weight matrix $R$: $\hat{A}= \hat{X}R\hat{X}$. There is a Sigmoid function to predict positive/negative labels from $\hat{A}$.

\begin{table*}[t]
\centering
\small
\begin{tabular}{lllllcllll}\toprule
  &  \multicolumn{4}{c}{NLP$\rightarrow$CV} &&  \multicolumn{4}{c}{NLP$\rightarrow$BIO} \\ \cline{2-5} \cline{7-10} 
 \textbf{Method}      & \textbf{F1}       & \textbf{Acc}      & \textbf{Pre}    & \textbf{Rec}   &  & \textbf{F1} &\textbf{Acc}        & \textbf{Pre}    & \textbf{Rec}   \\ 
 
\midrule
  \texttt{Baseline Models} & \multicolumn{9}{l}{} \\
   CLS + BERT   	& 0.4277 & 0.5480    &	0.5743	& 0.3419   & &  0.3930  & 0.6000            &  0.7481      &  0.2727      \\
   CLS + P2V   &   0.4881 	&\underline{0.5757}	&0.6106	& 0.4070 & & 0.2222 & 0.5333  & 0.6000 & 0.1364 \\
   VGAE + BERT \tiny{\cite{Li2019WhatSI}}  &0.5885 &	0.5477	 & 	0.5398 &	0.6488 &    &  0.6011   & 0.6091        &     0.6185 &	0.5909      \\
   VGAE + P2V \tiny{\cite{Li2019WhatSI}} & \underline{0.6202} & 0.5500  & 0.5368 & 0.7349 &  &     \underline{0.6177}   & \underline{0.6273}      &     0.6521 &	0.6091       \\
\midrule
  \textbf{\texttt{Proposed Method}} & \multicolumn{9}{l}{ }  \\
CD-VGAE + BERT    & 	0.6391 & 0.5593  & 	0.5441 & 	0.7884   &    &  0.6289 &  0.6273        &    0.6425	& 0.6364     \\
CD-VGAE + P2V     & \textbf{0.6754} & \textbf{0.5759}  & 0.5468 &	0.8837  &  & \textbf{0.6512} &   \textbf{0.6591}& 0.6667 &	0.6364\\ \midrule
\multicolumn{10}{l}{\texttt{Supervised Performance - Upper Bound}} \\
CLS + Node2vec \tiny{\cite{grover2016node2vec}}& 0.8172 & 0.8197  & 0.8223 & 0.8140   &   & 0.8060  &  0.7956         & 0.7547       & 0.8727       \\

\bottomrule
\end{tabular}
\caption{Evaluation results on two target domains. Underlined scores are the best among the baseline models. }
\label{tab:res}

\end{table*}

\section{Evaluation}

We evaluate on our new corpus LectureBankCD, treating the NLP domain as the source domain and transferring to the two new target domains: NLP$\rightarrow$CV and NLP$\rightarrow$BIO.  Consistent with \citet{kipf2016semi,Li2019WhatSI}, we randomly split the positive relations into 85\% training, 5\% validation, and 10\% testing. To account for imbalanced data, we randomly select negative relations such that the training set has the same number of positive and negative relations. We do the same for the validation and test sets. We report average scores over five different randomly seeded splits. 

To encode concepts and resources, we test BERT and P2V embeddings. For BERT, we applied a pre-trained version from Google\footnote{\url{https://github.com/google-research/bert}, (version with L = 12 and H = 768)}. We trained P2V using all the resource data. Both methods only require free-text for training and encoding.

\textbf{Baseline Models}
We concatenate the BERT/P2V embeddings of each pair of concepts and feed the result into a classifier (CLS + BERT and CLS + P2V). We train the classifier on the source domain only, then evaluate on the target domain. We report the best performance among Support Vector Machine, Logistic Regression, Gaussian Na\"ive Bayes, and Random Forest. In addition, we train the VGAE model \citet{Li2019WhatSI} on the source domain and test on the target domain, initializing the VGAE input with BERT and P2V embeddings separately (VGAE + BERT and VGAE + P2V). Given that GAE is structurally similar to VGAE, we leave this as a future work. Other graph-based methods including DeepWalk \cite{perozzi2014deepwalk} and Node2vec \cite{grover2016node2vec} are not applicable in this setting as both models require training edges from the target domain in order to generate node embeddings for target concepts. 

\begin{figure*}[t]
    \centering
    \includegraphics[width=16cm]{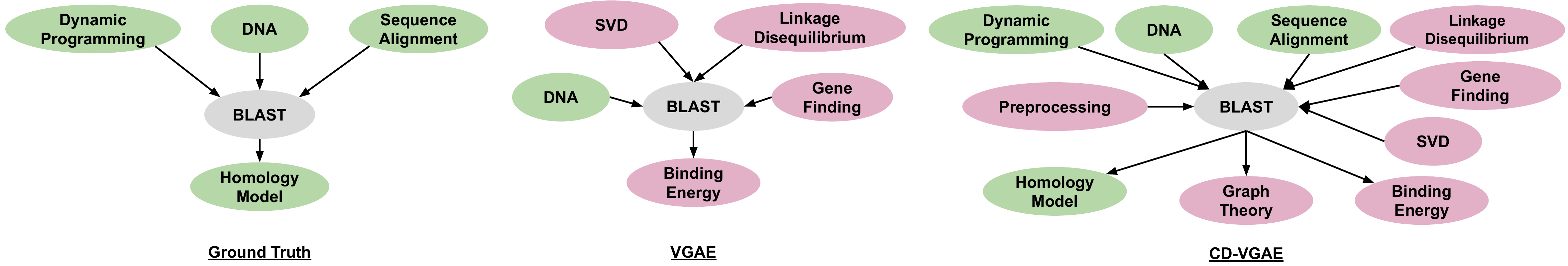}
    \caption{Case Study in BIO: direct neighbors of \textit{BLAST}, including successors and prerequisites, from the ground truth, VGAE, and our proposed CD-VGAE model. SVD stands for Singular Value Decomposition. Correct nodes are marked in blue, incorrect nodes are marked in red. (Best viewed in color!)}
    \label{fig:case}
\end{figure*}

\textbf{Proposed Method} We report results of our proposed model, CD-VGAE, initialized with BERT and P2V node embeddings separately. Consistent with the work from \citet{Li2019WhatSI} and \citet{li-etal-2020-r}, P2V embeddings yield better results than BERT embeddings in general. One possible reason for this difference is that BERT embeddings have a large number of dimensions, making it very easy to overfit. 
The two CLS models yield a negative result, with F1 worse than random guess. A possible reason is that treating concept pairs independently from the source domain may not be beneficial for the target domains. The VGAE models have a better performance when considering the concepts in a large graph. As shown in the table, our method performs better than the chosen baselines on both accuracy and F1 score, by incorporating information from domain neighbors. In particular, it yields much higher recall than all the baseline models. We provide further analysis in a later section.

\textbf{Upper Bound Performance} Finally, we conduct in-domain experiments on CV and BIO (supervised training and testing in the target domain), to show an upper bound for cross-domain performance. We test a variety of methods including traditional classifiers as well as graph-based approaches, including DeepWalk, Node2vec, and GraphSAGE \cite{graphsage17hamilton}. 

\section{Analysis}

Next, we conduct quantitative analysis and case studies on the target domain concept graphs recovered by our model (CD-VGAE+P2V) and two baseline models (CLS + P2V, VGAE + P2V), to take a closer look at the results.

\textbf{Quantitative Analysis} 
We first apply the three trained models to recover the concept graph in the CV domain. Compared to the ground truth with 871 positive relations, the baseline model predicts 527, VGAE predicts 963, and our model predicts 1,209. Similarly, in the BIO domain with 234 positive relations, the baseline model predicts only 128 positive edges, VGAE predicts 261, and our model predicts 303. Since our model tends to predict more positive edges, it has a higher recall. A higher recall is preferred in real-world applications as a system should not miss any relevant concepts when designing a user's study path.



\textbf{Concept Graph Recovery} 
We now provide case studies of the recovered concept graphs. In Table \ref{tab:case}, we show successors of the concept \textit{Image Processing} from the CV domain, i.e. concepts for which \textit{Image Processing} is a prerequisite. Both the baseline model and VGAE miss many successor concepts, whereas our model can recover a correct list without any missing concepts. 

We illustrate another case study from the BIO domain in Figure \ref{fig:case} using the concept \textit{BLAST} (short for \lq\lq basic local alignment search tool \rq \rq), an algorithm for comparing primary biological sequence information. In the ground truth, BLAST has three prerequisite concepts (\textit{Dynamic Programming}, \textit{DNA} and \textit{Sequence Alignment}), and one successor concept (\textit{Homology Model}). We observe that VGAE predicts only one prerequisite, \textit{DNA}, and misses all the others. In contrast, our model successfully includes all the ground truth relations, although it predicts some extra ones compared to VGAE. A closer look at the extra predictions reveals that these are still relevant topics, even though they are not direct prerequisites. For example, \textit{Sequence Alignment}, \textit{BLAST} and \textit{Graph Theory} are all associated with sequence analysis and share some common algorithms (i.e. De Bruijn Graph).

\begin{figure*}[th!]
    \centering
    \includegraphics[width=16cm]{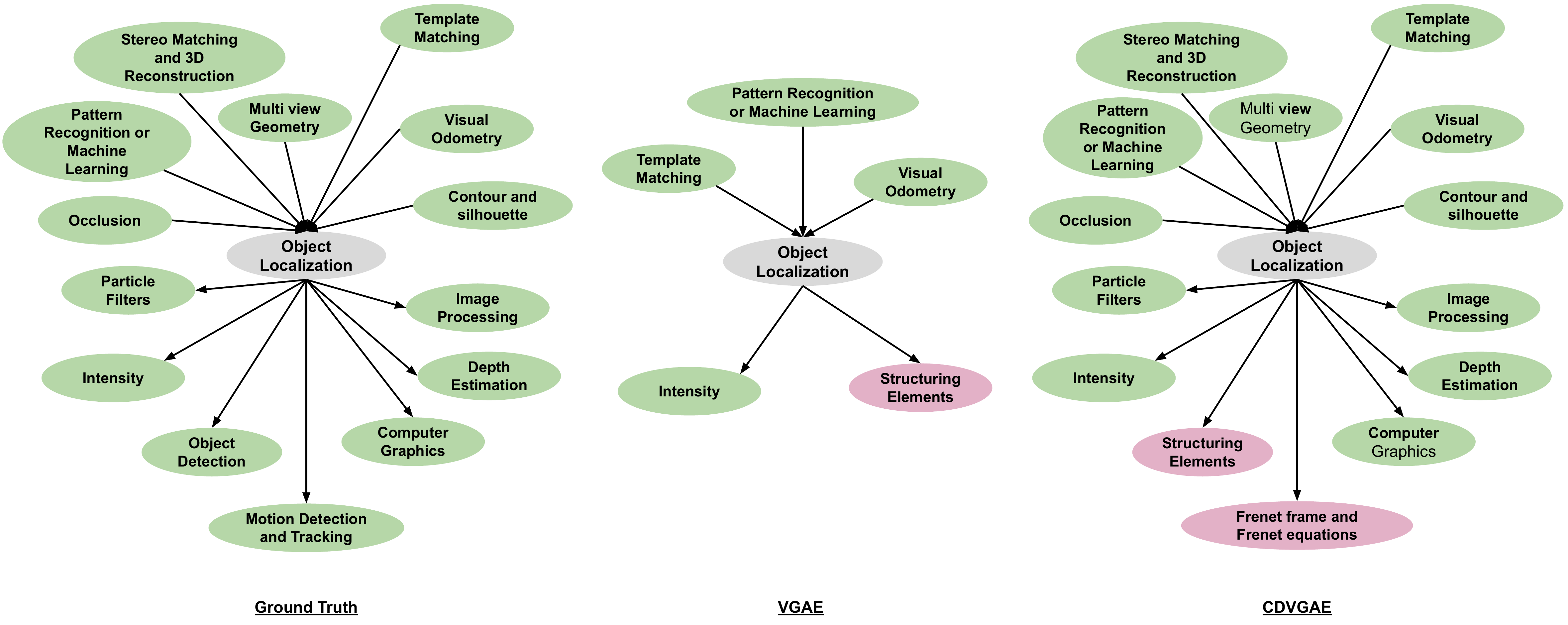}
    \caption{Case Study in CV: direct neighbors of \textit{Object Localization}.}
    \label{fig:cv}
\end{figure*}

We provide a case study in the CV domain, shown in Figure \ref{fig:cv}, by selecting concept node \textit{Object Localization}. The ground truth shows that it has 14 direct neighbors. The VGAE model only predicts five neighbors, while our model predicts more. Our model has two wrong predictions, but it gets 12 correct ones. In contrast, the VGAE model misses up to 10 neighbors, which is not acceptable in an application scenario of an educational platform leading students to miss very useful information. 

\begin{table}[t]
\centering
\scriptsize
\begin{tabular}{c|c} \toprule
\textbf{Base}                          & \textbf{VGAE}                        \\ \midrule
Image Representation          & Image Representation          \\
OCR & Computer graphics             \\
                              & Eye Tracking                  \\\toprule
\textbf{CD-VGAE}                         & \textbf{Ground Truth}                        \\\midrule
Video/Image augmentation  & Video/Image augmentation  \\
Image Representation          & Image Representation          \\
Face Detection                & Face detection                \\
Emotion Recognition           & Emotion Recognition           \\
Feature Extraction            & Feature Extraction            \\
Feature Learning              & Feature Learning              \\
OCR & OCR \\
Computer Graphics             & Computer Graphics             \\
Eye Tracking                & Eye Tracking                 \\\bottomrule
\end{tabular}
\caption{Successors of the concept \textit{Image Processing}, i.e. concepts for which \textit{Image Processing} is a prerequisite (OCR stands for Optical Character Recognition).}
\label{tab:case}
\end{table}

\section{Conclusion}
In this paper, we proposed the CD-VGAE model to solve the task of cross-domain prerequisite chain learning. Results show that our model outperforms previous unsupervised graph-based models by a large margin, especially regarding the F1 and recall score. Also, we created a new dataset that contains resources and concepts from two domains along with annotated prerequisite relations. 




\bibliographystyle{acl_natbib}
\bibliography{anthology,acl2021}

\newpage
\appendix

\section{Supervised Results}

\begin{table}[h!]
\centering
\small
\begin{tabular}{lcccc} \toprule
     \textbf{Method}            & \textbf{Acc}                              & \textbf{F1}                               & \textbf{Pre}    & \textbf{Rec}    \\ \midrule
GS+BERT & 0.7491                           & 0.7513                           & 0.7404 & 0.7628 \\
GS+P2V  & 0.7457                           & 0.7423                           & 0.7486 & 0.7372 \\
P2V              & 0.7642                           & 0.757                            & 0.7754 & 0.7395 \\
BERT             & 0.7572                           & 0.7495                           & 0.7677 & 0.7326 \\
DeepWalk         & 0.7988                           & 0.791                            & 0.8182 & 0.7674 \\
Node2vec         & \textbf{0.8197} & \textbf{0.8172} & 0.8223 & 0.8140 \\\bottomrule
\end{tabular}
\caption{Supervised evaluation results: CV$\rightarrow$CV. GS:GraphSAGE.}
\label{tab:res1}
\end{table}

\begin{table}[h!]
\centering
\small
\begin{tabular}{lcccc} \toprule
       \textbf{Method}            & \textbf{Acc}                              & \textbf{F1}                               & \textbf{Pre}    & \textbf{Rec}    \\ \midrule
GS+BERT & 0.7289                           & 0.7355                           & 0.7104 & 0.7727 \\
GS+P2V  & 0.7911                           & 0.7904                           & 0.7787 & 0.8091 \\
P2V              & 0.72                             & 0.7367                           & 0.6874 & 0.8091 \\
BERT             & 0.7067                           & 0.7189                           & 0.683  & 0.7727 \\
DeepWalk         & 0.7911                           & 0.8079                           & 0.7334 & 0.9091 \\
Node2vec         & \textbf{0.7956} & \textbf{0.8060} & 0.7547 & 0.8727 \\ \bottomrule
\end{tabular}
\caption{Supervised evaluation results: BIO$\rightarrow$BIO. GS:GraphSAGE. }
\label{tab:res2}
\end{table}



As a supplementary experiment, we present in-domain results in Table \ref{tab:res1}, \ref{tab:res2}: CV$\rightarrow$CV and BIO$\rightarrow$BIO, respectively. While we showed in the main paper that CLS + Node2vec yielded the best result, which serves as an upper bound on cross-domain performance, we additionally show our results from using other supervised methods:

\textbf{CLS + P2V/BERT} Encoding concept pairs with P2V/BERT, concatenating the embeddings of both concepts within each possible pair, and then train a binary classifier. We report the best performance among Support Vector Machine, Logistic Regression, Gaussian Na\"ive Bayes, and Random Forest.

\textbf{DeepWalk, Node2vec} DeepWalk \cite{perozzi2014deepwalk} randomly samples a node and traverses to a neighbor node randomly until it reaches a maximum length, updating the latent representation of each node after each \lq\lq walk \rq\rq to maximize the probability of each node's neighbors given a node's representation. Node2Vec \cite{grover2016node2vec} improves DeepWalk by providing the additional flexibility of placing weights on random walks. For both methods, we input the training prerequisite relations and get concept node embeddings. After generating embeddings for each concept in the target domain, for each concept pair, we concatenate the embeddings of both concepts and pass the concatenated representation into a classifier to predict the relation. Again, we report the best performance from the same four classifiers. 

\textbf{GraphSAGE + P2V/BERT} GraphSAGE \cite{graphsage17hamilton} is an inductive framework to generate node embeddings for unseen data by leveraging existing node features. We first treat it as a node embedding method, as done with DeepWalk and Node2vec. After concept node embedding is generated, we train a classifier to predict concept relations and report in-domain results. Besides, as an initial attempt, we investigate GraphSAGE for the out-of-domain setting. We assume that, because there are unseen topics when transferring to new domains, such an inductive method like GraphSAGE may fit in our scenario. However, we end up with negative results as the original GraphSAGE may not fit in to this specific application. We leave more investigation as future work. 

\end{document}